\documentclass[journal]{IEEEtran}
\usepackage{amsmath,amsfonts}
\usepackage{algorithmic}
\usepackage{algorithm}
\usepackage{array}
\usepackage{textcomp}
\usepackage{stfloats}
\usepackage{url}
\usepackage{verbatim}
\usepackage{graphicx}
\usepackage{cite}
\usepackage{subfigure}
\usepackage{amssymb}
\usepackage{color}
\usepackage{float}
\usepackage{mathtools}
\def\BibTeX{{\rm B\kern-.05em{\sc i\kern-.025em b}\kern-.08em
    T\kern-.1667em\lower.7ex\hbox{E}\kern-.125emX}}
\usepackage{balance}
\begin{document}
\title{Fast Stochastic Policy Gradient: Negative Momentum for Reinforcement Learning } 
\author{Haobin Zhang and Zhuang Yang~\IEEEmembership{Member,~IEEE}
\thanks{(Corresponding authors: Zhuang Yang). Zhuang Yang and Haobin Zhang is with the School of Computer Science and Technology, Soochow University, Suzhou 215006, China. (e-mail: zhuangyng@163.com; HaoBinZag@163.com)

This work was supported by grants from the National Natural Science Foundation of China under Grant 62302325. This work was also supported by the Natural Science Foundation of Jiangsu Province in China under Grant BK20230485 and by Project Funded by the Priority Academic Program Development of Jiangsu Higher Education Institutions.}}

\markboth{Journal of \LaTeX\ Class Files,~Vol.~18, No.~9, September~2024}%
{How to Use the IEEEtran \LaTeX \ Templates}

\maketitle

\begin{abstract}
Stochastic optimization algorithms, particularly stochastic policy gradient (SPG), report significant success in reinforcement learning (RL). Nevertheless, up to now, that how to speedily acquire an optimal solution for RL is still a challenge. To tackle this issue, this work develops a fast SPG algorithm from the perspective of utilizing a momentum, coined SPG-NM. Specifically, in SPG-NM, a novel type of the negative momentum (NM) technique is applied into the classical SPG algorithm. Different from the existing NM techniques, we have adopted a few hyper-parameters in our SPG-NM algorithm. Moreover, the computational complexity is nearly same as the modern SPG-type algorithms, e.g., accelerated policy gradient (APG), which equips SPG with Nesterov's accelerated gradient (NAG).  We evaluate the resulting algorithm on two classical tasks, bandit setting and Markov decision process (MDP). Numerical results in different tasks demonstrate faster convergence rate of the resulting algorithm by comparing state-of-the-art algorithms, which confirm the positive impact of NM in accelerating SPG for RL. Also, numerical experiments under different settings confirm the robustness of our SPG-NM algorithm for some certain crucial hyper-parameters, which ride the user feel free in practice.

\end{abstract}

\begin{IEEEkeywords}
reinforcement learning, stochastic policy gradient, accelerated gradient, negative momentum.
\end{IEEEkeywords}

\section{Introduction}
\IEEEPARstart{T}{he} policy gradient (PG) method is one of the most popular and effective reinforcement learning (RL) algorithms. In practice, the main goal of RL \cite{liu2023systematic,moerland2023model,shakya2023reinforcement,uprety2020reinforcement} is to explore how an agent studies a policy by the virtue of interacting with its environment to maximize the accumulative reward. PG \cite{sutton1999policy,bhandari2024global,daskalakis2020independent} is to parameterize the policy and optimize a target accumulated reward function through stochastic gradient ascent. More concretely, in order to learn the parameterized policy, the agent performs stochastic gradient ascent in the policy parameter space to maximize its expected return $J(\theta)$, where $\theta\in{\mathbb{R}^d}$ are parameters of the policy parameterization.

In difficult tasks involving tremendous state and action space, PG algorithms combined with the deep neural networks have demonstrated impressive empirical successes, thus making it more suitable for tackling RL tasks. Subsequently, the PG algorithm further was applied to the Actor-Critic \cite{konda1999actor} architecture, where we combine the PG method with the value function method. Such manipulation accelerated learning rate and improved stability of the classical PG algorithm in RL training. The oncoming of the Actor-Critic architecture modified most of the existing optimization algorithms, leading to a series of novel algorithms for RL, such as deterministic policy gradient algorithms (DPG)\cite{silver2014deterministic}, deep deterministic policy gradient (DDPG)\cite{lillicrap2015continuous}, trust region policy optimization (TRPO)\cite{schulman2015trust}, proximal policy optimization (PPO)\cite{schulman2017proximal}.

PG, based on Actor-Critic (AC) techniques, offer an attractive approach because arguably, unlike the Q-learning method, they are based on genuine conditions of optimality of the closed-loop policy. The stochastic policy gradient (SPG) method is a popular AC approach, and rather simple assumptions are required for the method to work, making it fairly robust to use. However, there is an obvious drawback in SPG, where SPG is more likely to get stuck in local optima or even fail to converge to the optimum sometimes. Therefore, it is still challenging that how to speedily obtain a suboptimal solution close to the optimum in RL.

There are several popular techniques to improve SPG, including but not limited to momentum, learning rate scheduling, adaptive learning rates, etc. Particularly, they improve SPG by accelerating stochastic gradient ascent strategy. With the help of these techniques, the plain PG method is low iteration complexity but converges faster (e.g., attaining a linear convergence rate). This work is more curious about the effect of the momentum in SPG. More recently, Recently, Chen \textit{et al.}\cite{chen2023accelerated} proposed APG, which uses Nesterov's momentum \cite{nesterov1983method} scheme to accelerate the convergence performance of PG in reinforcement learning, achieving rapid global convergence, and they formally establish that APG enjoys a $O(1/t^2)$ convergence rate under softmax policy parameterization. 
 
Motivated by recent development of optimization algorithms in RL, this paper proposes a novel SPG algorithm with fast and robust property, termed as SPG-NM, by applying negative momentum (NM) into SPG for RL. Negative momentum adjusts the update direction of the current step size according to the direction and magnitude of historical gradients, reducing oscillations during parameter updates and accumulating momentum in the gradient direction, thereby contributing to escaping local minima and approaching to the global optimum faster. 

For clarity and convenience, we further show that adding such momentum in SPG-like algorithms has a significant acceleration effect in the following three aspects, where there is also our main contribution of this work.

\begin{itemize}
\item[$\bullet$] We propose Stochastic Policy Gradient with Negative Momentum (SPG-NM), which leverages the NM scheme to accelerate the convergence performance of SPG for RL. We show that our SPG-NM algorithm enjoys the lower computational complexity as state-of-the-art SPG algorithms.

\item[$\bullet$] Numerical experiments on various RL tasks, bandit setting and MDP setting, show the superiority of our SPG-NM algorithm by comparing it with state-of-the-art SPG algorithms including PG, PG-HB, APG and PG-Adam.

\item[$\bullet$] Moreover, through further exploring the numerical performance of between our SPG-NM algorithm and the related algorithms on MDP problems under hard and uniform settings, we greatly show that SPG-NM attains a faster convergence rate and confirms the robustness of the resulting algorithms for some key hyper-parameters.

\end{itemize}

\section{Related Work}
\subsection{Policy Gradient} 
In modern reinforcement learning tasks, the value function and the state-value function is usually to be estimated, which means that agents must learn from the environment and make suitable decisions simultaneously. However, learning the state-value function is an extremely challenging task even invalid when the action space is large or even infinite. To deal with this issue, by parameterizing the policy directly, the REINFORCE algorithm \cite{williams1992simple} was developed that used gradient ascent to maximize the expected return. Particularly, the REINFORCE algorithm reported remarkable success in dealing with high-dimensional continuous action spaces. 

Subsequently, to more effectively estimate the value function, Silver \textit{et al.}\cite{silver2014deterministic} proposed a framework of DPG algorithms. Similarly, Lillicrap \textit{et al.} \cite{lillicrap2015continuous} proposed DDPG, where it applied the idea of deep Q-learning to continuous action tasks in reinforcement learning. Additionally, in order to optimize control policy with guaranteed, Schulman \textit{et al.} \cite{schulman2015trust} proposed TRPO through making several approximations to the theoretically-justified scheme, improving the robustness of each update by using a constraint on the KL divergence between the new and old policies in policy gradient. Additionally, as an extension of TRPO, PPO \cite{schulman2017proximal} ensured the stability of policy updates by introducing Clipped Surrogate Objective to limit the magnitude of policy updates.

To overcome the drawbacks of PG methods and value function methods, Actor-Critic methods were proposed by Konda \textit{et al.}, where the Actor learned the policy and the Critic evaluated the policy to guide the Actor's update direction. This combination allows this algorithm to leverage the advantages of policy optimization and value function estimation, thereby improving the numerical behavior and accelerating  the convergence speed of the algorithm. Inspired by the Actor-Critic method, there are also other variants of PG methods, including ACER \cite{wang2016sample}, A3C \cite{mnih2016asynchronous} and its synchronous variant A2C, full name (ACKTR)\cite{wu2017scalable}, and SAC\cite{haarnoja2018soft}. Recently, the global convergence of standard PG methods in various settings has been proven by Agarwal \textit{et al.}\cite{agarwal2021theory}. In addition, (Fazel \textit{et al.}\cite{fazel2018global}, 2018; Liu \textit{et al.}\cite{liu2020improved}, 2020; Wang \textit{et al.}\cite{wang2021global}, 2021; Xiao\cite{xiao2022convergence}, 2022) have theoretically analyzed several PG methods under different policy parameterizations and established global convergence guarantees for these methods.

\subsection{Accelerated Gradient} 
The heavy ball method \cite{polyak1964some} is well known for its simplicity, practicality, and high efficiency. Subsequently, Yurii Nesterov invented the Nesterov Accelerated Gradient (NAG) algorithm for smooth functions \cite{nesterov1983method}. Then, he proposed a more widely applicable NAG-type method in \cite{nesterov1988general}. More specifically, the NAG algorithm has been theoretically proven to achieve a convergence rate of $O(1/t^2)$ for strongly convex and smooth convex functions, surpassing the limited convergence rate of $O(1/t)$ provided by gradient descent algorithms. Obviously, anyone can safely conclude that NAG converge faster than traditional gradient descent methods. Interestingly, the convergence behavior of the Nesterov method can be explained from the perspectives of ordinary differential equations \cite{lessard2016analysis}, geometry \cite{allen2017linear}, and game theory \cite{lan2018optimal}. Lately, NAG-type algorithms were applied into non-smooth objective functions and composite functions, leading to various variants of accelerated gradient algorithms.

Moreover, Beck \textit{et al.} \cite{beck2009fast} introduced the Fast Iterative Shrinkage Thresholding Algorithm (FISTA), combining proximal operators with accelerated gradient methods to better handle optimization problems with sparsity or constraints by performing proximal projection operations at each iteration. Nesterov \cite{nesterov2004introductory} combined coordinate descent with accelerated gradient methods, updating only one coordinate axis at each iteration to decrease computational complexity and accelerate convergence speed. Ghadimi\textit{et al.}\cite{ghadimi2016accelerated} proposed a generalization of the NAG method to solve general nonlinear optimization problems. Li \textit{et al.}\cite{li2015convolutional} introduced the APG-type algorithm, which effectively solves general non-convex and non-smooth problems by introducing a monitor with sufficient descent properties, providing convergence guarantees for stable points of non-convex problems. Recently, Yang \cite{yang2024sarah} introduced a momentum term to the stochastic recursive gradient algorithm (SARAH), presented a novel variance reduced stochastic gradient algorithm (SARAH-M), this algorithm outperforms state-of-the-art stochastic optimization algorithms under a specific condition.

Furthermore, Chen \textit{et al.} \cite{chen2023accelerated} proposed a new APG method, which used the NAG scheme to accelerate the convergence performance of PG in reinforcement learning, achieving a rapid global convergence. The above list of works is not exhaustive but serves as a brief overview of accelerated gradient methods. This paper introduces SPG-NM, a novel RL method that combines accelerated gradient methods with PG methods, introducing NM to adjust each update, thereby achieving a fast convergence rate.

\section{Method}
In this section, we first introduce MDP in subsection \ref{sub-m-1}. Then, we present our proposed algorithm, SPG-NM, in subsection \ref{sub-m-2} which integrates negative momentum with gradient-based reinforcement learning algorithms.

\subsection{Markov Decision Process}
\label{sub-m-1}
Markov Decision Process (MDP) is a mathematical framework, often describing a dynamic system with randomness and decision-making. MDP is commonly employed in reinforcement learning to model the interaction between an agent and its environment. Here is a detailed explanation of MDP. Concretely, $\mathcal{M} = (\mathcal{S}, \mathcal{A}, \mathcal{P}, r, \gamma, \rho)$ is determined by: (i) the set of all possible states that the system can be in, denoted as state pace $\mathcal{S}$, (ii) the set of all possible actions that the agent can take, denoted as action space $\mathcal{A}$, (iii) the agent receives the immediate reward for taking an action in a particular state. Usually denoted as reward function $\mathcal{S} \times \mathcal{A} \to \mathbb{R}$, indicating the reward obtained from taking action $a$ in state$s$, (iv) the probability of transitioning from state $s$ to state $s'$ by taking action $a$. Typically denoted as transition probability $\mathcal{P}(s'\rvert s,a)$, (v) a discount factor $\gamma \in [0, 1)$, and (vi) an initial state distribution $\rho \in \Delta(\mathcal{S})$. Without loss of generality, we presume the one step reward lies in the $[0,1]$ interval. Given a policy $\pi : \mathcal{S} \to \Delta(\mathcal{A})$, the value of state $s$ under $\pi$ is defined as
\begin{align}
\label{eq:state_value_function}
V^{\pi}(s) = \mathbb{E}\bigg[\sum_{t=0}^{\infty} \gamma^{t} r(s_{t}, a_{t})\bigg\vert \pi, s_{0}=s\bigg].
\end{align}

The agent starts from an initial state $s_{0}$ at time $t = 0$. At each time step $t$, the agent selects an action $a_{t}$ based on the current state $s_{t}$ and executes $a_t$. After performing the action $a_{t}$, the agent, varying from state $s_{t}$ to state $s_{t+1}$ based on the transition probability $\mathcal{P}(s_{t+1}\rvert s_{t},a_{t})$, receives a reward $\mathbb{R}(s_{t},a_{t})$. Value functions are used to evaluate the long-term return for the agent in a state or state-action pair. More specifically, value functions in MDPs often refer into value functions $V(s)$ and action value functions $Q(s,a)$. On the other side, the goal of MDP is to find an optimal policy $\pi^*$ of the optimal value functions, $V^{\pi^*}(s)$ or $Q^{\pi^*}(s,a)$, by maximizing the agent's long-term return, where the optimal policy satisfies the Bellman Optimality Equation:
\begin{align}
\label{eq:bellman_optimality_equation-1}
V^{\pi^*}(s) & = \mathop{\max}_{{a}\in\mathcal{A}}\bigg[{\mathbb{R}(s,a) + \gamma \sum_{s'}{P}(s'\rvert s,a)V^{\pi}(s')}\bigg], \\
or \notag\\
\label{eq:bellman_optimality_equation-2}
Q^{\pi^*}(s,a) & = \mathbb{R}(s,a) + \gamma \sum_{s'}{P}(s'\rvert s,a)\mathop{\max}_{a'}Q^{\pi}(s',a').
\end{align}

Solving the above Bellman Optimality Equation, \eqref{eq:bellman_optimality_equation-1} or \eqref{eq:bellman_optimality_equation-2}, we can reach a stable state, $V^{\pi^*}(s)$ or $Q^{\pi^*}(s,a)$, where in the stable state the value function no longer changes anymore. In other words, at the stable state, the value function corresponds to the optimal state value function for the given policy.

\subsection{SPG with Negative Momentum}
\label{sub-m-2}
In order to follow the idea of this work, this part states from introducing PG, NAG and Katyusha.
The target of PG method is modeling and optimizing the policy at the same time. Concretely, the policy is usually modeled with a parameterized function respecting to $\theta$ and $\mathnormal{\pi}_{\theta}$($\mathnormal{a}$$\vert$$\mathnormal{s}$). PG method explores the space of parameters to make an optimal policy $\pi^*$, thereby obtaining the best reward. More specifically, the pseudo code of PG is shown in Algorithm 1, developed in \cite{sutton1999policy}.
\begin{algorithm} [!hptb]
	\caption{Policy Gradient Method (PG)}
	\label{algorithm:PG}
	\begin{algorithmic}
\STATE \textbf{Input}:
Learning rate $\eta^{(t)} > 0$.
\STATE \textbf{Initialize}:
$\theta^{(0)}(s,a)$ for all $(s,a)$.
\FOR{$t = 1$ to $T$} 
    \STATE
    \begin{align} 
    \theta^{(t)} \leftarrow \theta^{(t-1)} + \eta \nabla_{\theta}{V^{\pi_{\theta}}(\mu)} \Big\rvert_{\theta = \theta^{(t)}} \
    \end{align}
\ENDFOR
	\end{algorithmic}
\end{algorithm}

Further, the NAG-type iterative scheme is presented in  Algorithm \ref{algorithm:NAG}. As seen from Algorithm \ref{algorithm:NAG}, NAG adopts the current and historical information to update the solution simultaneously, making NAG converge fast in practice. 

\begin{algorithm} [!hptb]
	\caption{Nesterov’s Accelerated Gradient (NAG)\cite{su2016differential}}
	\label{algorithm:NAG}
	\begin{algorithmic}
\STATE \textbf{Input}:
Learning rate $s = \frac{1}{L}$, where $L$ is the Lipschitz constant of the function $f$.
\STATE \textbf{Initialize}:
${x}^{(0)}$ and ${y}^{(0)} = {x}^{(0)}$.
\FOR{$t = 1$ to $T$} 
    \STATE
    \begin{align} 
    & {x}^{(t)} = {y}^{(t-1)} - s \nabla f({y}^{(t-1)}) \\
    & {y}^{(t)} = {x}^{(t)} + \frac{t-1}{t+2}({x}^{(t)}-{x}^{(t-1)})
    \end{align}
\ENDFOR
	\end{algorithmic}
\end{algorithm}

As pointed by many studies, in the stochastic setting, NAG algorithm is prone to acquiring error accumulation and struggles to achieve accelerated convergence in general. By introducing a ``negative momentum", Katyusha \cite{allen2018katyusha} successfully solve this issue in NAG with noisy gradients, where the Katyusha method usually performs the following updates for $k = 1$ to $N$ :
\begin{align}
\label{eq:Katyusha}
x_{k+1} & \leftarrow {\tau}_{1}z_k + {\tau}_{2}{\Tilde{x}} + (1 - {\tau}_{1} -  {\tau}_{2}){y}_{k} \\
\Tilde{\nabla}_{k+1} & \leftarrow \nabla f(\Tilde{x}) + \nabla f_i(x_{k+1}) - \nabla f_i(\Tilde{x}) \\
y_{k+1} & \leftarrow x_{k+1} -  \frac{1}{3L}\Tilde{\nabla}_{k+1} \\
z_{k+1} & \leftarrow z_k - \alpha \Tilde{\nabla}_{k+1}
\end{align}

In the formula above, $\Tilde{\nabla}_{k+1}$ is the gradient estimator which is utilized for ensuring that its variance approaches to zero as $k$ grows, $\Tilde{x}$ is a snapshot point which is updated every m iterations, ${\tau}_{1}$ and ${\tau}_{2}$ are two momentum parameters (${\tau}_{1}, {\tau}_{2} \in [0,1]$), $\alpha$ is equal to $\frac{1}{3{\tau}_{1}L}$.

Motivated by above mentioned algorithms, we, thereby, propose our  algorithm, SPG with NM, and present the pseudo code of our algorithm.
\begin{algorithm} [!ht]
	\caption{SPG with Negative Momentum(SPG-NM)}
	\label{algorithm:PG-NM}
	\begin{algorithmic}
\STATE \textbf{Input}:
Learning rate $\eta^{(t)} > 0$, $\lambda > 0$.
\STATE \textbf{Initialize}:
$\theta^{(0)} \in \mathbb{R}^{|\mathcal{S}||\mathcal{A}|}$, $\omega^{(0)} = \theta^{(0)}$.
\FOR{$t = 1$ to $T$} 
    \STATE
    \begin{align} 
    \theta^{(t)} &\leftarrow \omega^{(t-1)} + \eta^{(t)} \nabla_{\theta}{V^{\pi_{\theta}}(\mu)} \Big\rvert_{\theta = \omega^{(t-1)}}
    \label{algorithm:NM1}\\
    \varphi^{(t)} &\leftarrow \lambda\theta^{(t)} + (1-\lambda)(\theta^{(t)}-\theta^{(t-1)}) 
    \label{algorithm:eq2}\\
    \omega^{(t)} &\leftarrow
    \begin{cases}
        \varphi^{(t)}, & \text{if } V^{\pi_{\varphi}^{(t)}}(\mu) \ge V^{\pi_{\theta}^{(t)}}(\mu), \\
        \theta^{(t)}, & \text{otherwise.}
    \end{cases}
    \label{algorithm:eq3}
    \end{align}
\ENDFOR
	\end{algorithmic}
\end{algorithm}

{\bf{Remark:}} A few explanations for SPG-NM is provided here:

\begin{itemize}
\item[i)] In SPG-NM, the momentum term ($\theta^{(t)}-\theta^{(t-1)}$) ensures that $\varphi^{(t+1)}$ and $\varphi^{(t)}$ are not too far apart, and uses ($1-\lambda$) to adjust the magnitude of momentum. Thus, The momentum is like the ``magnet", which ensures that the solution of each iteration does not far away from the solution of last iteration, thus keeping each update sufficiently accurate. Generally, the SPG-NM method can be viewed as incorporating the NM term into the classical PG method,  which counteracts a portion of the positive momentum generated in the early iterations and therefore accelerates the convergence speed of the PG-like algorithm. 

\item[ii)] The momentum coefficient $\lambda$ in our SPG-NM algorithm needs to be adjusted. In our experiments, we show that selecting the momentum coefficient $\lambda$ from $10^{3}$ to $10^{5}$ is enough to ensure the better performance of SPG-NM. Moreover, we empirically show that a larger momentum coefficient retains more historical information during updates, resulting in a fast convergence rate of the algorithm. But the side effect of adopting a larger  $\lambda$ is making the algorithm oscillate near local optimal and generating  a decrease of the value function in reinforcement learning. On the contrary, a smaller momentum coefficient reduces oscillations but may result in slower convergence. 

\item[iii)] Additionally, in order to further improve the performance of RL in practical problems, we introduce a novel updating rule for the solution $\omega^{(t)}$ in SPG-NM, where we only update the current parameters (a.k.a. $\omega^{(t)}=\varphi^{(t)}$ if the current $V^{\pi_{\varphi}^{(t)}}(\mu)$ is greater than the $V^{\pi_{\theta}^{(t)}}(\mu)$ in the previous iteration. Otherwise, we still use the previous solution, i.e., $\omega^{(t)}=\theta^{(t)}$.
\end{itemize}

\section{Experiment}
In this section, we empirically validate the effectiveness of SPG-NM by performing experiments on a 3-armed bandit and MDP with 5 states and 5 actions. Note that without otherwise specified, the value of $\lambda$ is set to be $\lambda=1000$. Then, we explore the numerical performance of our SPG-NM algorithm with the momentum coefficient $\lambda$ by executing numerical experiments on MDP. Besides, the sub-optimality gap of different algorithms is also provided to greatly confirm the efficacy of NM in enhancing the conventional PG method. More specifically, to show the superiority of our SPG-NM algorithm, PG \cite{sutton1999policy}, PG-HB \cite{su2016differential}, PG-Adam \cite{kingma2014adam}, and APG \cite{chen2023accelerated} are provided as benchmark algorithms.

\subsection{Bandit} \label{sub-ban-I} We run a 3-action bandit experiment with actions $\mathcal A$ = [$\mathnormal{a}^{*}$, $\mathnormal{a}_{2}$, $\mathnormal{a}_{3}$], where the corresponding rewards are $\mathnormal{r}$ = [$\mathnormal{r}$($\mathnormal{a}^{*}$), $\mathnormal{r}$($\mathnormal{a}_{2}$), $\mathnormal{r}$($\mathnormal{a}_{3}$)] = [${1.0}$, ${0.99}$, ${0}$]. We initialize the policy parameters with both a uniform initialization ($\theta^{(0)}$ = [${0}$, ${0}$, ${0}$], $\pi^{(0)}$ = [${1/3}$, ${1/3}$, ${1/3}$]) and a hard initialization ($\theta^{(0)}$ = [${1}$, ${3}$, ${5}$], $\pi^{(0)}$ = [${0.01588}$, ${0.11731}$, ${0.86681}$]).

\begin{figure}[htbp]
\centering
\subfigure[SPG-NM]
{
    \begin{minipage}[b]{.9\linewidth}
        \centering
        \includegraphics[scale=0.29]{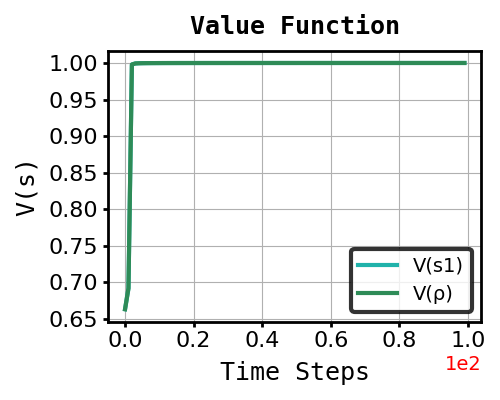}
    \end{minipage}
}
\subfigure[APG]
{
    \begin{minipage}[b]{.45\linewidth}
        \centering
        \includegraphics[scale=0.29]{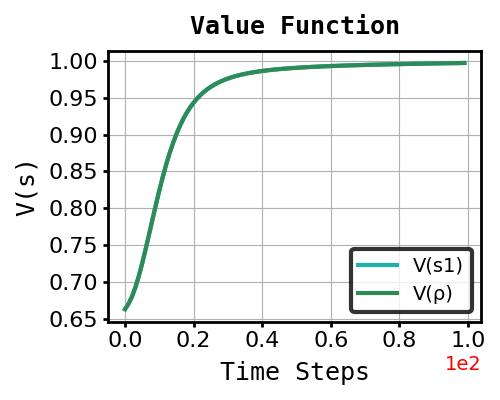}  
    \end{minipage}
}
\subfigure[PG-Adam]
{
    \begin{minipage}[b]{.45\linewidth}
        \centering
        \includegraphics[scale=0.29]{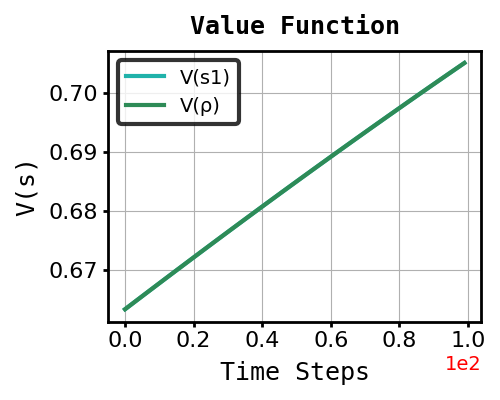}
    \end{minipage}
}
\subfigure[PG]
{
    \begin{minipage}[b]{.45\linewidth}
        \centering
        \includegraphics[scale=0.29]{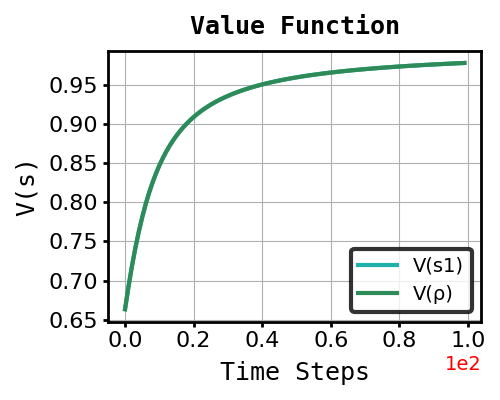}
    \end{minipage}
}
\subfigure[PG-HB]
{
    \begin{minipage}[b]{.45\linewidth}
        \centering
        \includegraphics[scale=0.29]{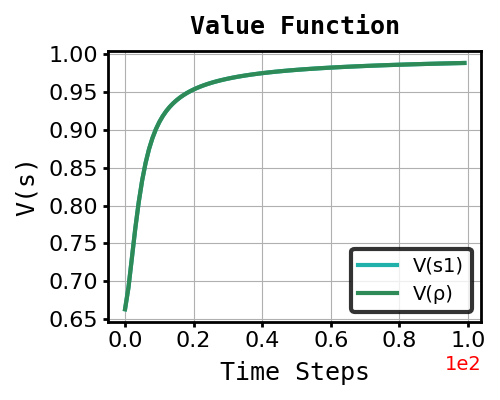} 
    \end{minipage}
}
\caption{A comparison between the performance of five different algorithms under bandit, and uniform policy initialization:(a)-(e) show the value functions of them.}
\label{fig:bandit-uniform}
\end{figure}

\begin{figure}[htbp]
\centering
\subfigure[SPG-NM]
{
    \begin{minipage}[b]{.9\linewidth}
        \centering
        \includegraphics[scale=0.29]{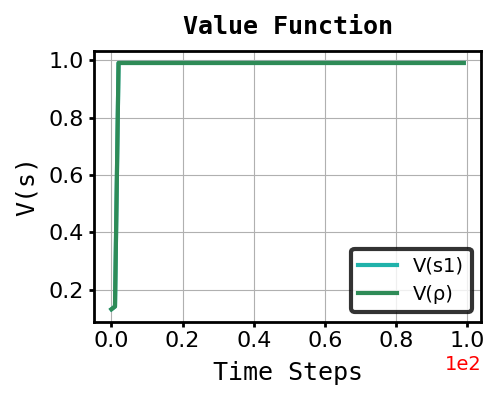}
    \end{minipage}
}
\subfigure[APG]
{
    \begin{minipage}[b]{.45\linewidth}
        \centering
        \includegraphics[scale=0.29]{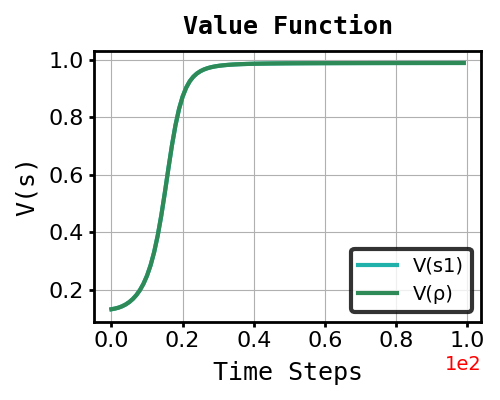}
    \end{minipage}
}
\subfigure[PG-Adam]
{
    \begin{minipage}[b]{.45\linewidth}
        \centering
        \includegraphics[scale=0.29]{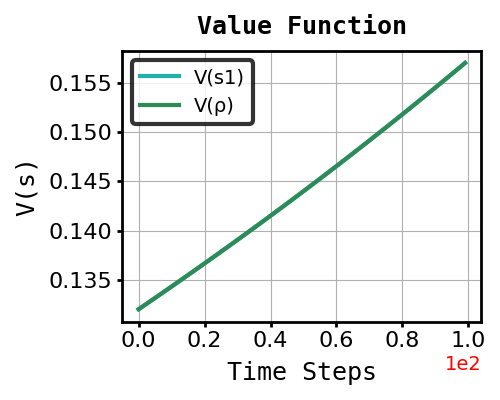}  
    \end{minipage}
}
\subfigure[PG]
{
    \begin{minipage}[b]{.45\linewidth}
        \centering
        \includegraphics[scale=0.29]{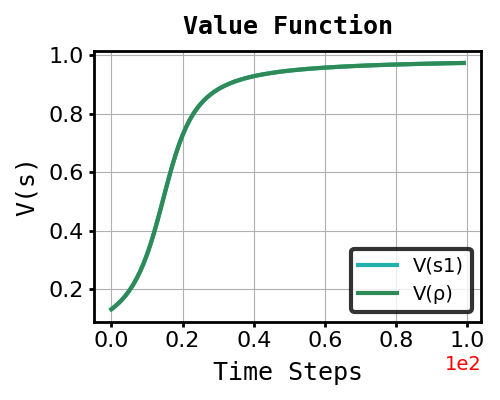}
    \end{minipage}
}
\subfigure[PG-HB]
{
    \begin{minipage}[b]{.45\linewidth}
        \centering
        \includegraphics[scale=0.29]{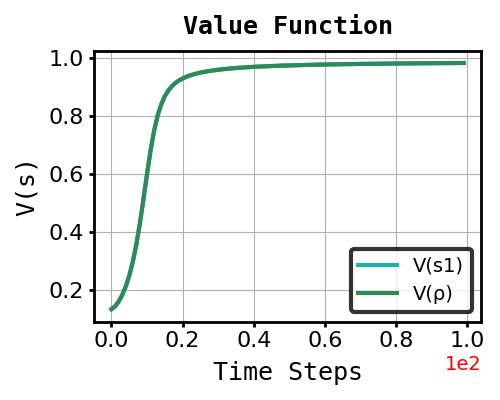}  
    \end{minipage}
}
\caption{A comparison between the performance of five different algorithms under bandit, and hard policy initialization:(a)-(e) show the value functions of them.}
\label{fig:bandit-hard}
\end{figure}

Concretely, we start from performing a 3-armed bandit experiment with both a uniform initialization and a hard initialization and show the results in Fig. \ref{fig:bandit-uniform} and Fig. \ref{fig:bandit-hard}. The goal of this experiment is to explore the convergence speed of the algorithm by showing the case of different algorithms how to approximate maximize cumulative rewards. In Fig. \ref{fig:bandit-uniform} and Fig. \ref{fig:bandit-hard}, the $x$ label denotes the time that the agent interacts with its environment and the $y$ label denotes the long-term cumulative reward that the agent obtains.

Fig. \ref{fig:bandit-uniform} shows the performance of different algorithms on bandit with uniform policy initialization. As observed by Fig. \ref{fig:bandit-uniform}, our SPG-NM algorithm converges faster than  PG, PG-HB, APG, and PG-Adam under bandit with uniform policy initialization. In addition, Fig. \ref{fig:bandit-uniform} demonstrates that PG-Adam fails to fast converge to the stationary point.

Further, Fig. \ref{fig:bandit-hard} explores the numerical performance of various algorithms by conducting numerical experiments on bandit with hard policy initialization. We emphasize that in contrast to a bandit with uniform policy initialization, a bandit with hard policy initialization means that the optimal action has the minimum initial probability. It is pointed out by Fig. \ref{fig:bandit-hard} that SPG-NM still maintains a faster convergence speed than the comparative results (a.k.a. PG, PG-HB, APG, and PG-Adam). 

Both Fig. \ref{fig:bandit-uniform} and Fig. \ref{fig:bandit-hard} indicate that APG do not show their advantage under different tasks. Moreover, Fig. \ref{fig:bandit-uniform} and Fig. \ref{fig:bandit-hard} demonstrate that PG-Adam that equips the HBM-type technique into PG perform worse under different tasks. In contrast, our SPG-NM algorithm achieves a better performance, which further confirms the efficacy of the NM technique in improving the classical SPG algorithm. Importantly, the results in Fig. \ref{fig:bandit-uniform} and Fig. \ref{fig:bandit-hard} implies that our SPG-NM algorithm is more suitable for RL problems such as online decision making.

The above experiments are performed on a simple task that is  a 3-armed bandit experiment. To show the much promising of our SPG-NM algorithm, in the following, we perform experiments on a difficult task, MDP, which is also widely used to test the efficacy of the algorithm for RL.

\subsection{MDP} We perform an experiment on MDP with 5 states and 5 actions under the initial state distribution $\rho$ = [$0.3$, $0.2$, $0.1$, $0.15$, $0.25$]. The reward function $\mathnormal{r}$($\mathnormal{s,a}$) = (($\mathnormal{s}_{1}$, $\mathnormal{a}_{1}$),($\mathnormal{s}_{2}$, $\mathnormal{a}_{2}$), ($\mathnormal{s}_{3}$, $\mathnormal{a}_{3}$), ($\mathnormal{s}_{4}$, $\mathnormal{a}_{4}$), ($\mathnormal{s}_{5}$, $\mathnormal{a}_{5}$)) = [[$1.0$, $0.8$, $0.6$, $0.7$, $0.4$], [$0.5$, $0.3$, $0.1$, $1.0$, $0.6$], [$0.6$, $0.9$, $0.8$, $0.7$, $1.0$], [$0.1$, $0.2$, $0.6$, $0.7$, $0.4$], [$0.8$, $0.4$, $0.6$, $0.2$, $0.9$]]. The initial policy parameters are both a uniform initialization (A 5x5 matrix full of zeros) and a hard initialization ($\theta_{s,a}^{(0)}$ = (($\mathnormal{s}_{1}$, $\mathnormal{a}_{1}$),($\mathnormal{s}_{2}$, $\mathnormal{a}_{2}$), ($\mathnormal{s}_{3}$, $\mathnormal{a}_{3}$), ($\mathnormal{s}_{4}$, $\mathnormal{a}_{4}$), ($\mathnormal{s}_{5}$, $\mathnormal{a}_{5}$)) = [[$1$, $2$, $3$, $4$, $5$], [$3$, $4$, $5$, $1$, $2$], [$5$, $2$, $3$, $4$, $1$], [$5$, $4$, $2$, $1$, $3$], [$2$, $4$, $3$, $5$, $1$]]). In addition, there are five transition probability matrices ($\mathnormal{P}$($\mathnormal{s}$$\vert$$\mathnormal{s}_{0}$, $\mathnormal{a}$) = (($\mathnormal{s}_{1}$, $\mathnormal{a}_{1}$),($\mathnormal{s}_{2}$, $\mathnormal{a}_{2}$), ($\mathnormal{s}_{3}$, $\mathnormal{a}_{3}$), ($\mathnormal{s}_{4}$, $\mathnormal{a}_{4}$), ($\mathnormal{s}_{5}$, $\mathnormal{a}_{5}$)) = [[$0.1$, $0.6$, $0.5$, $0.4$, $0.2$], [$0.5$, $0.1$, $0.1$, $0.3$, $0.1$], [$0.1$, $0.1$, $0.1$, $0.1$, $0.1$], [$0.2$, $0.1$, $0.2$, $0.1$, $0.1$], [$0.1$, $0.1$, $0.1$, $0.1$, $0.5$]], $\mathnormal{P}$($\mathnormal{s}$$\vert$$\mathnormal{s}_{1}$, $\mathnormal{a}$) = (($\mathnormal{s}_{1}$, $\mathnormal{a}_{1}$),($\mathnormal{s}_{2}$, $\mathnormal{a}_{2}$), ($\mathnormal{s}_{3}$, $\mathnormal{a}_{3}$), ($\mathnormal{s}_{4}$, $\mathnormal{a}_{4}$), ($\mathnormal{s}_{5}$, $\mathnormal{a}_{5}$)) = [[$0.1$, $0.4$, $0.1$, $0.4$, $0.2$], [$0.5$, $0.1$, $0.4$, $0.1$, $0.2$], [$0.2$, $0.2$, $0.3$, $0.1$, $0.2$], [$0.1$, $0.2$, $0.1$, $0.1$, $0.2$], [$0.1$, $0.1$, $0.1$, $0.3$, $0.2$]], $\mathnormal{P}$($\mathnormal{s}$$\vert$$\mathnormal{s}_{2}$, $\mathnormal{a}$) = (($\mathnormal{s}_{1}$, $\mathnormal{a}_{1}$),($\mathnormal{s}_{2}$, $\mathnormal{a}_{2}$), ($\mathnormal{s}_{3}$, $\mathnormal{a}_{3}$), ($\mathnormal{s}_{4}$, $\mathnormal{a}_{4}$), ($\mathnormal{s}_{5}$, $\mathnormal{a}_{5}$))= [[$0.6$, $0.2$, $0.3$, $0.1$, $0.2$], [$0.1$, $0.4$, $0.3$, $0.4$, $0.1$], [$0.1$, $0.1$, $0.2$, $0.3$, $0.1$], [$0.1$, $0.2$, $0.1$, $0.1$, $0.1$], [$0.1$, $0.1$, $0.1$, $0.1$, $0.5$]], $\mathnormal{P}$($\mathnormal{s}$$\vert$$\mathnormal{s}_{3}$, $\mathnormal{a}$) = (($\mathnormal{s}_{1}$, $\mathnormal{a}_{1}$),($\mathnormal{s}_{2}$, $\mathnormal{a}_{2}$), ($\mathnormal{s}_{3}$, $\mathnormal{a}_{3}$), ($\mathnormal{s}_{4}$, $\mathnormal{a}_{4}$), ($\mathnormal{s}_{5}$, $\mathnormal{a}_{5}$)) = [[$0.6$, $0.1$, $0.2$, $0.4$, $0.5$], [$0.1$, $0.5$, $0.1$, $0.3$, $0.1$], [$0.1$, $0.1$, $0.1$, $0.1$, $0.1$], [$0.1$, $0.2$, $0.1$, $0.1$, $0.2$], [$0.1$, $0.1$, $0.5$, $0.1$, $0.1$]], $\mathnormal{P}$($\mathnormal{s}$$\vert$$\mathnormal{s}_{4}$, $\mathnormal{a}$) = (($\mathnormal{s}_{1}$, $\mathnormal{a}_{1}$),($\mathnormal{s}_{2}$, $\mathnormal{a}_{2}$), ($\mathnormal{s}_{3}$, $\mathnormal{a}_{3}$), ($\mathnormal{s}_{4}$, $\mathnormal{a}_{4}$), ($\mathnormal{s}_{5}$, $\mathnormal{a}_{5}$)) = [[$0.2$, $0.4$, $0.4$, $0.1$, $0.2$], [$0.2$, $0.1$, $0.1$, $0.4$, $0.5$], [$0.2$, $0.2$, $0.1$, $0.2$, $0.1$], [$0.2$, $0.2$, $0.3$, $0.1$, $0.1$], [$0.2$, $0.1$, $0.1$, $0.2$, $0.1$]]).

\begin{figure}[htbp]
\centering
\subfigure[SPG-NM]
{
    \begin{minipage}[b]{.9\linewidth}
        \centering
        \includegraphics[scale=0.29]{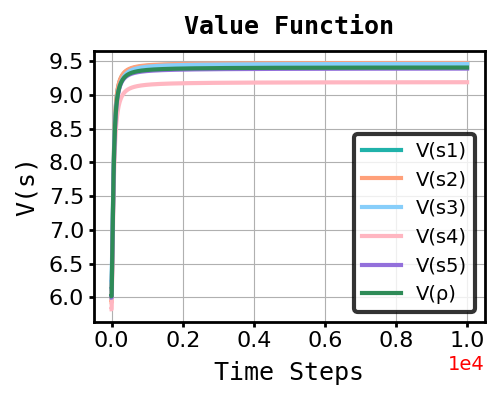}
    \end{minipage}
}
\subfigure[APG]
{
    \begin{minipage}[b]{.45\linewidth}
        \centering
        \includegraphics[scale=0.29]{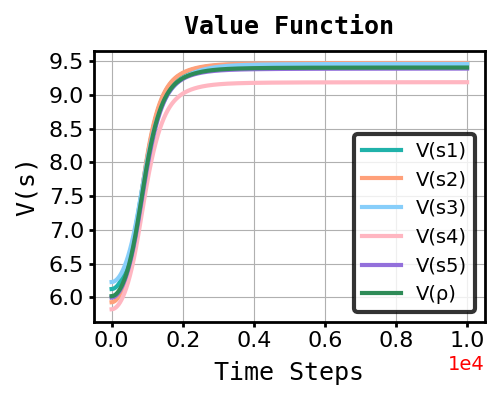}
    \end{minipage}
}
\subfigure[PG-Adam]
{
    \begin{minipage}[b]{.45\linewidth}
        \centering
        \includegraphics[scale=0.29]{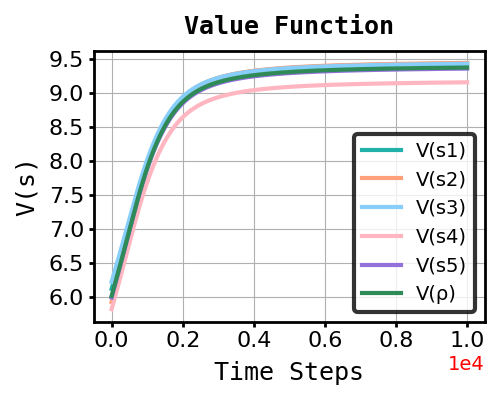}  
    \end{minipage}
}
\subfigure[PG]
{
    \begin{minipage}[b]{.45\linewidth}
        \centering
        \includegraphics[scale=0.29]{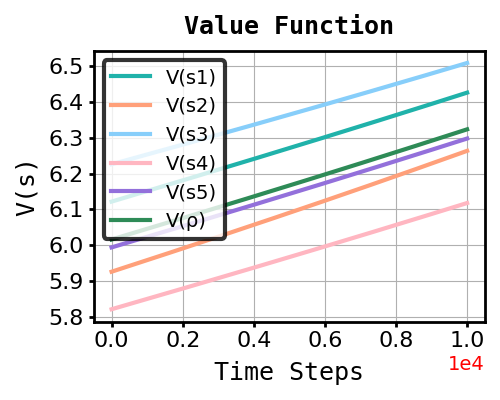}
    \end{minipage}
}
\subfigure[PG-HB]
{
    \begin{minipage}[b]{.45\linewidth}
        \centering
        \includegraphics[scale=0.29]{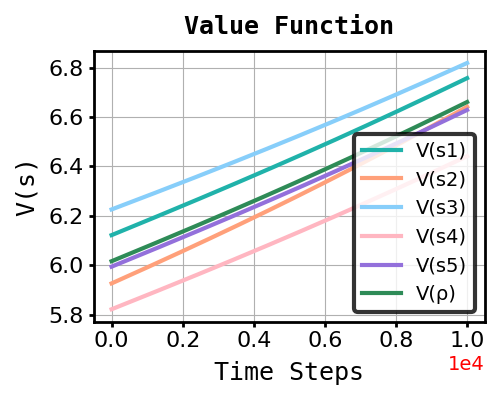}  
    \end{minipage}
}
\caption{A comparison between the performance of five different algorithms under MDP with 5 states, 5 actions, and uniform policy initialization:(a)-(e) show the per-state value functions of them.}
\label{fig:MDP-uniform}
\end{figure}

\begin{figure}[htbp]
\centering
\subfigure[SPG-NM]
{
    \begin{minipage}[b]{.9\linewidth}
        \centering
        \includegraphics[scale=0.29]{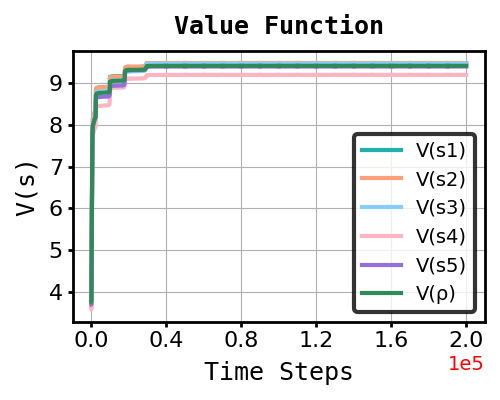}
    \end{minipage}
}
\subfigure[APG]
{
    \begin{minipage}[b]{.45\linewidth}
        \centering
        \includegraphics[scale=0.29]{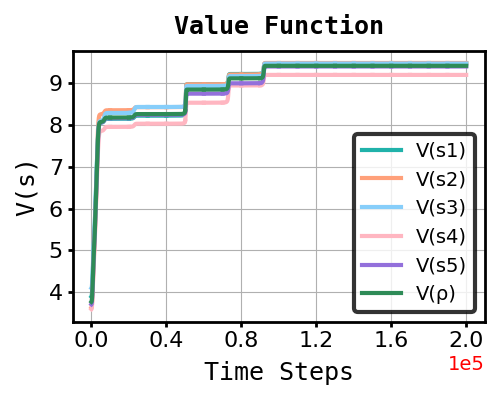}
    \end{minipage}
}
\subfigure[PG-Adam]
{
    \begin{minipage}[b]{.45\linewidth}
        \centering
        \includegraphics[scale=0.29]{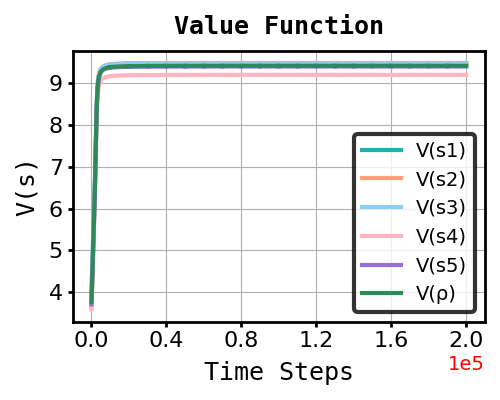}  
    \end{minipage}
}
\subfigure[PG]
{
    \begin{minipage}[b]{.45\linewidth}
        \centering
        \includegraphics[scale=0.29]{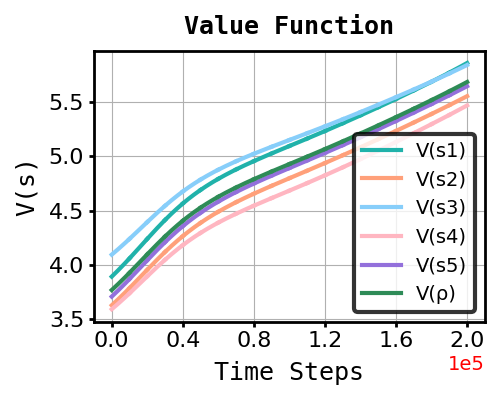}
    \end{minipage}
}
\subfigure[PG-HB]
{
    \begin{minipage}[b]{.45\linewidth}
        \centering
        \includegraphics[scale=0.29]{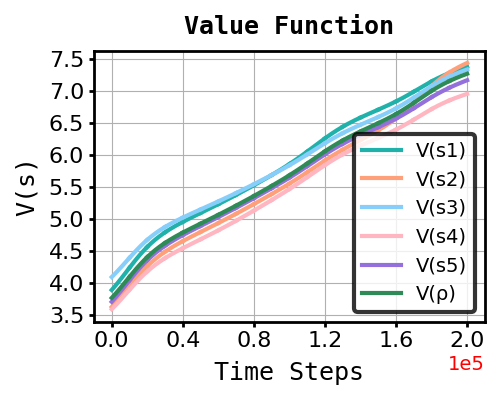}  
    \end{minipage}
}
\caption{A comparison between the performance of five different algorithms under MDP with 5 states, 5 actions, and hard policy initialization:(a)-(e) show the per-state value functions of them.}
\label{fig:MDP-hard}
\end{figure}

We emphasize that the main goal of this segment is to show the efficacy of the algorithms by exploring the cases of convergence behavior and cumulative rewards at the same time. More specifically, the numerical results of  MDP with 5 states, 5 actions under both a uniform initialization and a hard initialization are plotted in  Fig. \ref{fig:MDP-uniform} and Fig. \ref{fig:MDP-hard}. Here, in all figures, the $x$ label and the $y$ label have the same meaning as figures in subsection \ref{sub-ban-I}.

Fig. \ref{fig:MDP-uniform} demonstrates the performance of various algorithms on MDP under a uniform initialization. The results in Fig. \ref{fig:MDP-uniform} clearly show that SPG-NM attains a faster convergence speed than PG, PG-HB, PG-Adam, and APG. Additionally, Fig. \ref{fig:MDP-uniform} demonstrates that the convergence speed of PG and PG-HB is extremely slow.

Followed by, Fig. \ref{fig:MDP-hard} displays the performance of different algorithms in terms of convergence speed and cumulative rewards on MDP under a hard policy initialization. Compared with MDP under a uniform policy initialization, the optimal policy under a hard policy initialization is more difficult. Observe Fig. \ref{fig:MDP-hard}, we have that our SPG-NM algorithm converges a slightly more slowly than PG-Adam but faster than  PG, PG-HB, and APG. This performance in Fig. \ref{fig:MDP-hard} demonstrates that SPG-NM still has a faster convergence speed in a difficult environment.

More generally, although Fig. \ref{fig:MDP-hard} show that the convergence speed of SPG-NM is slower than PG-Adam on MDP under a hard policy initialization, PG-Adam shows extremely slow convergence speed  under bandit, both a uniform policy initialization and a hard policy initialization, which has been shown in Fig. \ref{fig:bandit-uniform} and Fig. \ref{fig:bandit-hard}. Importantly, SPG-NM confirms its superiority on both bandit experiments and MDP experiments, showing the much promising of our SPG-NM algorithm in diverse practical applications. In addition, compared to SPG-NM, APG, and PG Adam, the standard PG algorithms (PG and PG-HB) perform worse in more complex RL tasks.

Algorithm \ref{algorithm:PG-NM} shows that SPG-NM may highly depend on the hyper-parameter $\lambda$. However, in the above experiments, we do not specially selecting this crucial parameter for different tasks. To better comprehend the effect of the hyper-parameter, $\lambda$, in SPG-NM (Algorithm \ref{algorithm:PG-NM}), we will show the performance of SPG-NM (Algorithm \ref{algorithm:PG-NM}) with different $\lambda$ in subsection \ref{sub-l-III}.

\subsection{Choice of the hyper-parameter $\lambda$}
\label{sub-l-III}
Here, we test the impact of $\lambda$ in  SPG-NM (Algorithm \ref{algorithm:PG-NM}) by performing experiments on MDP, where the parameter settings are similar to above section. For $\lambda$, we start with ${10}^{3}$ and increase the value of $\lambda$ by a factor of ten on MDP experiments gradually.

\begin{figure}[htbp]
\centering
\subfigure[$\lambda = 1000$]
{
    \begin{minipage}[b]{.45\linewidth}
        \centering
        \includegraphics[scale=0.29]{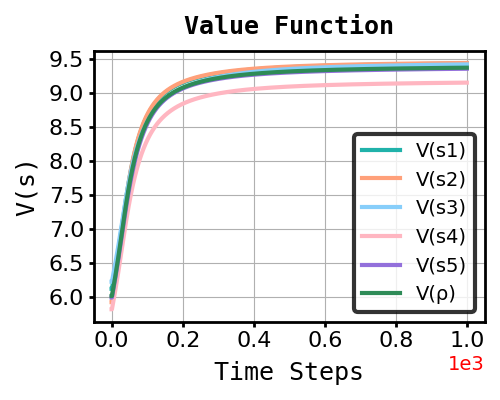}
    \end{minipage}
}
\subfigure[$\lambda = 10000$]
{
    \begin{minipage}[b]{.45\linewidth}
        \centering
        \includegraphics[scale=0.29]{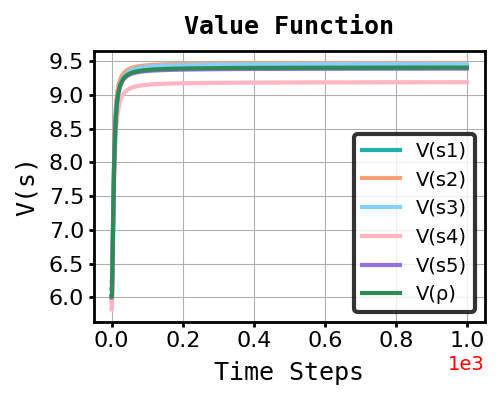}
    \end{minipage}
}
\subfigure[$\lambda = 100000$]
{
    \begin{minipage}[b]{.45\linewidth}
        \centering
        \includegraphics[scale=0.29]{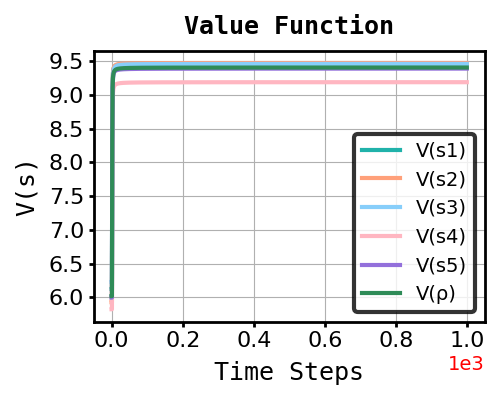}  
    \end{minipage}
}
\subfigure[$\lambda = 1000000$]
{
    \begin{minipage}[b]{.45\linewidth}
        \centering
        \includegraphics[scale=0.29]{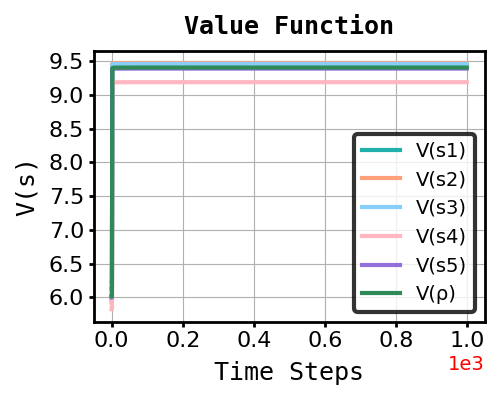}
    \end{minipage}
}
\caption{A comparison of different $\lambda$ under MDP with 5 states, 5 actions, and a uniform policy initialization.}
\label{fig:parameter-MDP-uniform}
\end{figure}

Fig. \ref{fig:parameter-MDP-uniform} shows the performance of SPG-NM (Algorithm \ref{algorithm:PG-NM}) under different selections for $\lambda$. In Fig. \ref{fig:parameter-MDP-uniform}, we observe that no matter what value $\lambda$ takes, our algorithm performs fast convergence speed and attains high value function in a simple environment. Moreover, as the value of $\lambda$ increases, the convergence speed becomes faster. The result in Fig. \ref{fig:parameter-MDP-uniform} makes us conclude that SPG-NM (Algorithm \ref{algorithm:PG-NM}) is insensitive to $\lambda$.

\begin{figure}[htbp]
\centering
\subfigure[$\lambda = 1000$]
{
    \begin{minipage}[b]{.45\linewidth}
        \centering
        \includegraphics[scale=0.29]{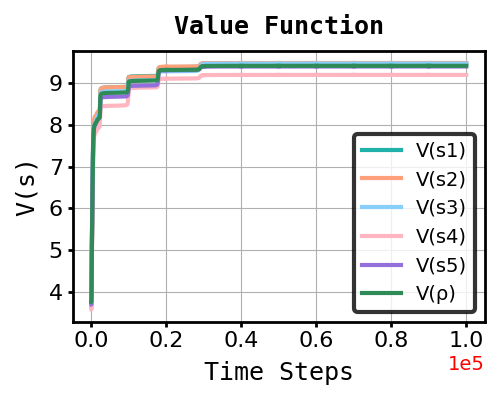}
    \end{minipage}
}
\subfigure[$\lambda = 10000$]
{
    \begin{minipage}[b]{.45\linewidth}
        \centering
        \includegraphics[scale=0.29]{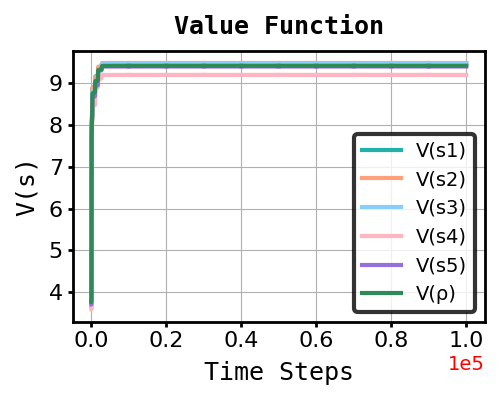}
    \end{minipage}
}
\subfigure[$\lambda = 100000$]
{
    \begin{minipage}[b]{.45\linewidth}
        \centering
        \includegraphics[scale=0.29]{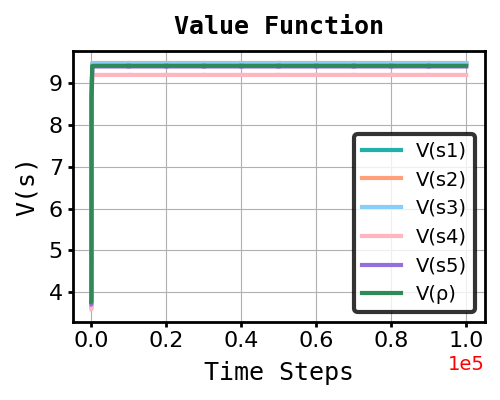}  
    \end{minipage}
}
\subfigure[$\lambda = 1000000$]
{
    \begin{minipage}[b]{.45\linewidth}
        \centering
        \includegraphics[scale=0.29]{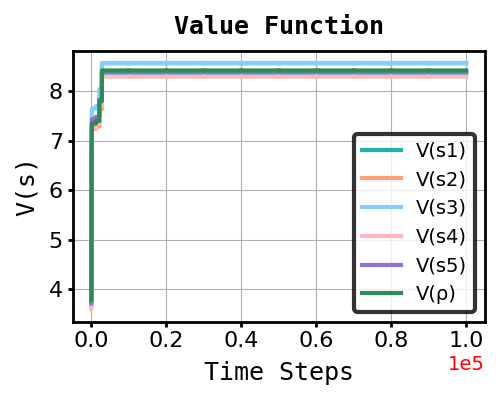}
    \end{minipage}
}
\caption{A comparison of different $\lambda$ under MDP with 5 states, 5 actions, and a hard policy initialization.}
\label{fig:parameter-MDP-hard}
\end{figure}

Fig. \ref{fig:parameter-MDP-hard} investigates the numerical performance of SPG-NM (Algorithm \ref{algorithm:PG-NM}) under different values of $\lambda$ by performing numerical experiments on MDP under a hard policy initialization. Obviously, we observe from Fig. \ref{fig:parameter-MDP-hard} that the performance of our algorithm becomes worse under all $\lambda$ on a more challenging environments, compared to MDP under a uniform policy initialization. Additionally, Fig. \ref{fig:parameter-MDP-hard} explains that the convergence speed of our algorithm becomes faster with the increases of $\lambda$ when the value of $\lambda$ is below ${10}^{6}$. However, PG-NM's performance descends sharply when the magnitude of $\lambda$ arrives ${10}^{6}$. Practically, as the value of $\lambda$ continues to increase, the convergence speed becomes slower and the value function decreases further. 

In summary, the changes of $\lambda$ have a slight impact on SPG-NM for simple problems, but our algorithm requires adjustment of the parameter $\lambda$ to achieve better performance on some difficult RL tasks.

\subsection{Sub-Optimality Gap}

\begin{figure}[htbp]
\centering
\subfigure[MDP-uniform]
{
    \begin{minipage}[b]{.45\linewidth}
        \centering
        \includegraphics[scale=0.29]{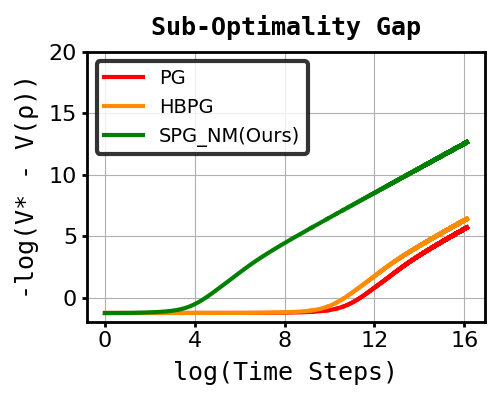}
    \end{minipage}
}
\subfigure[MDP-hard]
{
    \begin{minipage}[b]{.45\linewidth}
        \centering
        \includegraphics[scale=0.29]{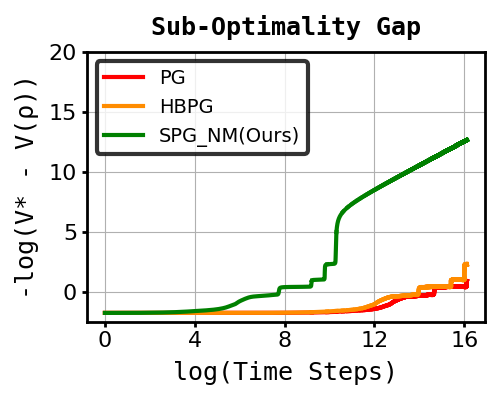}
    \end{minipage}
}
\caption{A comparison between PG, PG-HB, and SPG-NM under MDP with uniform policy initialization and hard policy initialization.}
\label{fig:sub-optimality gap}
\end{figure}

To significantly validate the efficacy of NM in improving the classical SPG algorithm, we show the sub-optimality gap among our SPG-NM algorithm, PG and HBPG.  Particularly, experiments are also conducted on MDP and the results are plotted in Fig. \ref{fig:sub-optimality gap}.

Fig. \ref{fig:sub-optimality gap} shows the performance on sub-optimality gap of SPG-NM, PG, and PG-HB on MDP with both a uniform policy initialization and a hard policy initialization. The results in Fig. \ref{fig:sub-optimality gap} demonstrate that the sub-optimality gap of SPG-NM is much smaller than PG and PG-HB. Furthermore, this result indicates that SPG-NM obtain better optimal policy in contrast to PG and PG-HB. Summarily, the integration of momentum term and standard PG algorithms greatly improves the performance of the standard PG algorithms. 

\section{Conclusion}

This paper proposed SPG-NM as a novel optimization technique for RL problems. Our SPG-NM algorithm enjoyed the lower computational complexity as state-of-the-art SPG algorithms, but converged faster than state-of-the-art SPG algorithms. Concretely, in SPG-NM, we incorporated NM into canonical SPG algorithm to improve the performance of the original SPG algorithm. Furthermore, numerical experiments on various RL tasks, bandit setting and MDP setting, show the advantage of our SPG-NM algorithm through comparing state-of-the-art SPG algorithms, including PG, PG-HB, APG and PG-Adam. Additionally, we studied the effect of crucial hyper-parameter $\lambda$ in our SPG-NM algorithm under different tasks and showed the robustness of our SPG-NM algorithm to $\lambda$.

In this work, we select a constant learning rate (a.k.a. $\eta_k=\eta$) for SPG-NM, which is widely adopted in the existing PG-like algorithms, e.g., PG, PG-HB, APG, etc. However, we believe that an appropriated rule of updating the learning rate will lead to a better performance of PG-like algorithms. As a consequence, future work will develop several update rules of the learning rate. In addition, future work will show much promise of our SPG-NM in other different practical applications, including but not limited to robot path planning problem, online decision, recommendation systems, etc. 

\bibliographystyle{ieeetr}
\bibliography{ref}

\begin{thebibliography}{10}

\bibitem{liu2023systematic}
W.~Liu, M.~Hua, Z.~Deng, Z.~Meng, Y.~Huang, C.~Hu, S.~Song, L.~Gao, C.~Liu, B.~Shuai, {\em et~al.}, ``A systematic survey of control techniques and applications in connected and automated vehicles,'' {\em IEEE Internet of Things Journal}, 2023.

\bibitem{moerland2023model}
T.~M. Moerland, J.~Broekens, A.~Plaat, C.~M. Jonker, {\em et~al.}, ``Model-based reinforcement learning: A survey,'' {\em Foundations and Trends{\textregistered} in Machine Learning}, vol.~16, no.~1, pp.~1--118, 2023.

\bibitem{shakya2023reinforcement}
A.~K. Shakya, G.~Pillai, and S.~Chakrabarty, ``Reinforcement learning algorithms: A brief survey,'' {\em Expert Systems with Applications}, p.~120495, 2023.

\bibitem{uprety2020reinforcement}
A.~Uprety and D.~B. Rawat, ``Reinforcement learning for iot security: A comprehensive survey,'' {\em IEEE Internet of Things Journal}, vol.~8, no.~11, pp.~8693--8706, 2020.

\bibitem{sutton1999policy}
R.~S. Sutton, D.~McAllester, S.~Singh, and Y.~Mansour, ``Policy gradient methods for reinforcement learning with function approximation,'' {\em Advances in neural information processing systems}, vol.~12, 1999.

\bibitem{bhandari2024global}
J.~Bhandari and D.~Russo, ``Global optimality guarantees for policy gradient methods,'' {\em Operations Research}, 2024.

\bibitem{daskalakis2020independent}
C.~Daskalakis, D.~J. Foster, and N.~Golowich, ``Independent policy gradient methods for competitive reinforcement learning,'' {\em Advances in neural information processing systems}, vol.~33, pp.~5527--5540, 2020.

\bibitem{konda1999actor}
V.~Konda and J.~Tsitsiklis, ``Actor-critic algorithms,'' {\em Advances in neural information processing systems}, vol.~12, 1999.

\bibitem{silver2014deterministic}
D.~Silver, G.~Lever, N.~Heess, T.~Degris, D.~Wierstra, and M.~Riedmiller, ``Deterministic policy gradient algorithms,'' in {\em International conference on machine learning}, pp.~387--395, Pmlr, 2014.

\bibitem{lillicrap2015continuous}
T.~P. Lillicrap, J.~J. Hunt, A.~Pritzel, N.~Heess, T.~Erez, Y.~Tassa, D.~Silver, and D.~Wierstra, ``Continuous control with deep reinforcement learning,'' {\em arXiv preprint arXiv:1509.02971}, 2015.

\bibitem{schulman2015trust}
J.~Schulman, S.~Levine, P.~Abbeel, M.~Jordan, and P.~Moritz, ``Trust region policy optimization,'' in {\em International conference on machine learning}, pp.~1889--1897, PMLR, 2015.

\bibitem{schulman2017proximal}
J.~Schulman, F.~Wolski, P.~Dhariwal, A.~Radford, and O.~Klimov, ``Proximal policy optimization algorithms,'' {\em arXiv preprint arXiv:1707.06347}, 2017.

\bibitem{chen2023accelerated}
Y.-J. Chen, N.-C. Huang, and P.-C. Hsieh, ``Accelerated policy gradient: On the nesterov momentum for reinforcement learning,'' in {\em ICML Workshop on New Frontiers in Learning, Control, and Dynamical Systems}, 2023.

\bibitem{nesterov1983method}
Y.~Nesterov, ``A method of solving a convex programming problem with convergence rate o (1/k** 2),'' {\em Doklady Akademii Nauk SSSR}, vol.~269, no.~3, p.~543, 1983.

\bibitem{williams1992simple}
R.~J. Williams, ``Simple statistical gradient-following algorithms for connectionist reinforcement learning,'' {\em Machine learning}, vol.~8, pp.~229--256, 1992.

\bibitem{wang2016sample}
Z.~Wang, V.~Bapst, N.~Heess, V.~Mnih, R.~Munos, K.~Kavukcuoglu, and N.~de~Freitas, ``Sample efficient actor-critic with experience replay,'' in {\em International Conference on Learning Representations}, 2016.

\bibitem{mnih2016asynchronous}
V.~Mnih, A.~P. Badia, M.~Mirza, A.~Graves, T.~Lillicrap, T.~Harley, D.~Silver, and K.~Kavukcuoglu, ``Asynchronous methods for deep reinforcement learning,'' in {\em International conference on machine learning}, pp.~1928--1937, PMLR, 2016.

\bibitem{wu2017scalable}
Y.~Wu, E.~Mansimov, R.~B. Grosse, S.~Liao, and J.~Ba, ``Scalable trust-region method for deep reinforcement learning using kronecker-factored approximation,'' {\em Advances in neural information processing systems}, vol.~30, 2017.

\bibitem{haarnoja2018soft}
T.~Haarnoja, A.~Zhou, P.~Abbeel, and S.~Levine, ``Soft actor-critic: Off-policy maximum entropy deep reinforcement learning with a stochastic actor,'' in {\em International conference on machine learning}, pp.~1861--1870, PMLR, 2018.

\bibitem{agarwal2021theory}
A.~Agarwal, S.~M. Kakade, J.~D. Lee, and G.~Mahajan, ``On the theory of policy gradient methods: Optimality, approximation, and distribution shift,'' {\em Journal of Machine Learning Research}, vol.~22, no.~98, pp.~1--76, 2021.

\bibitem{fazel2018global}
M.~Fazel, R.~Ge, S.~Kakade, and M.~Mesbahi, ``Global convergence of policy gradient methods for the linear quadratic regulator,'' in {\em International conference on machine learning}, pp.~1467--1476, PMLR, 2018.

\bibitem{liu2020improved}
Y.~Liu, K.~Zhang, T.~Basar, and W.~Yin, ``An improved analysis of (variance-reduced) policy gradient and natural policy gradient methods,'' {\em Advances in Neural Information Processing Systems}, vol.~33, pp.~7624--7636, 2020.

\bibitem{wang2021global}
W.~Wang, J.~Han, Z.~Yang, and Z.~Wang, ``Global convergence of policy gradient for linear-quadratic mean-field control/game in continuous time,'' in {\em International Conference on Machine Learning}, pp.~10772--10782, PMLR, 2021.

\bibitem{xiao2022convergence}
L.~Xiao, ``On the convergence rates of policy gradient methods,'' {\em Journal of Machine Learning Research}, vol.~23, no.~282, pp.~1--36, 2022.

\bibitem{polyak1964some}
B.~T. Polyak, ``Some methods of speeding up the convergence of iteration methods,'' {\em Zhurnal Vychislitel'noi Matematiki i Matematicheskoi Fiziki}, vol.~4, no.~5, pp.~791--803, 1964.

\bibitem{nesterov1988general}
Y.~Nesterov and A.~Nemirovsky, ``A general approach to polynomial-time algorithms design for convex programming,'' {\em Report, Central Economical and Mathematical Institute, USSR Academy of Sciences, Moscow}, vol.~182, 1988.

\bibitem{lessard2016analysis}
L.~Lessard, B.~Recht, and A.~Packard, ``Analysis and design of optimization algorithms via integral quadratic constraints,'' {\em SIAM Journal on Optimization}, vol.~26, no.~1, pp.~57--95, 2016.

\bibitem{allen2017linear}
Z.~Allen-Zhu and L.~Orecchia, ``Linear coupling: An ultimate unification of gradient and mirror descent,'' in {\em 8th Innovations in Theoretical Computer Science Conference (ITCS 2017)}, Schloss-Dagstuhl-Leibniz Zentrum f{\"u}r Informatik, 2017.

\bibitem{lan2018optimal}
G.~Lan and Y.~Zhou, ``An optimal randomized incremental gradient method,'' {\em Mathematical programming}, vol.~171, pp.~167--215, 2018.

\bibitem{beck2009fast}
A.~Beck and M.~Teboulle, ``A fast iterative shrinkage-thresholding algorithm for linear inverse problems,'' {\em SIAM journal on imaging sciences}, vol.~2, no.~1, pp.~183--202, 2009.

\bibitem{nesterov2004introductory}
Y.~Nesterov, ``Introductory lectures on convex optimization,'' {\em Applied Optimization}, vol.~87, 2004.

\bibitem{ghadimi2016accelerated}
S.~Ghadimi and G.~Lan, ``Accelerated gradient methods for nonconvex nonlinear and stochastic programming,'' {\em Mathematical Programming}, vol.~156, no.~1, pp.~59--99, 2016.

\bibitem{li2015convolutional}
H.~Li, Z.~Lin, X.~Shen, J.~Brandt, and G.~Hua, ``A convolutional neural network cascade for face detection,'' in {\em Proceedings of the IEEE conference on computer vision and pattern recognition}, pp.~5325--5334, 2015.

\bibitem{yang2024sarah}
Z.~Yang, ``Sarah-m: A fast stochastic recursive gradient descent algorithm via momentum,'' {\em Expert Systems with Applications}, vol.~238, p.~122295, 2024.

\bibitem{su2016differential}
W.~Su, S.~Boyd, and E.~J. Candes, ``A differential equation for modeling nesterov's accelerated gradient method: Theory and insights,'' {\em Journal of Machine Learning Research}, vol.~17, no.~153, pp.~1--43, 2016.

\bibitem{allen2018katyusha}
Z.~Allen-Zhu, ``Katyusha: The first direct acceleration of stochastic gradient methods,'' {\em Journal of Machine Learning Research}, vol.~18, no.~221, pp.~1--51, 2018.

\bibitem{kingma2014adam}
D.~Kingma, ``Adam: a method for stochastic optimization,'' in {\em Int Conf Learn Represent}, 2014.

\end{thebibliography}

\begin{IEEEbiography}[{\includegraphics[width=1in,height=1.25in,clip,keepaspectratio]{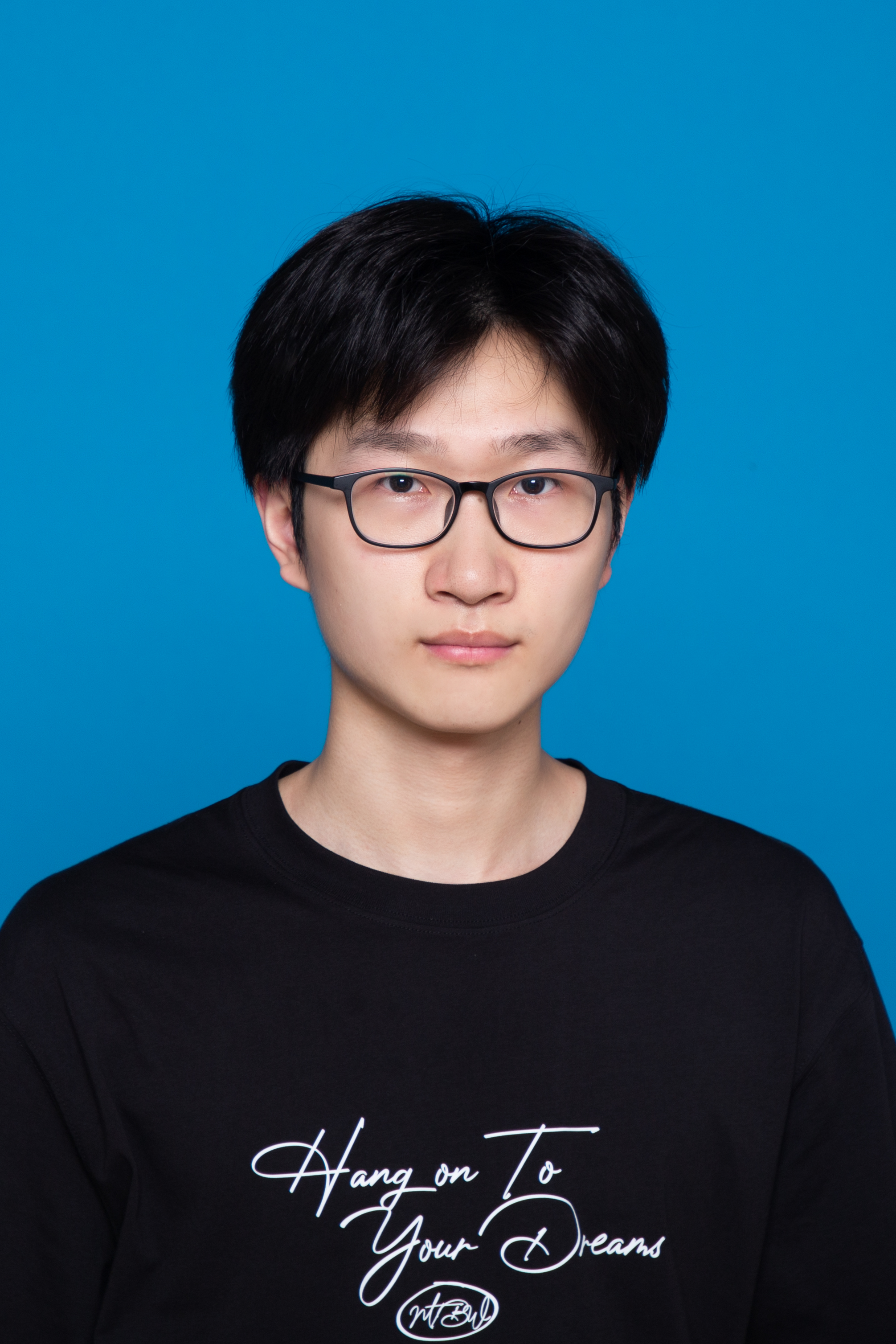}}]
{Hao Bin Zhang} received B.Eng. degree in School of Software Engineering from Jinling University, Nanjing, China in 2022. He is currently a postgraduate student in the School of Computer Science and Technology, Soochow University, Suzhou, China. His current research interest includes reinforcement learning, dynamical systems, data mining,  and computer vision. His recent work has focused on high-performance optimization algorithms for reinforcement learning and other related areas.
\end{IEEEbiography}

\begin{IEEEbiography}[{\includegraphics[width=1in,height=1.25in,clip,keepaspectratio]{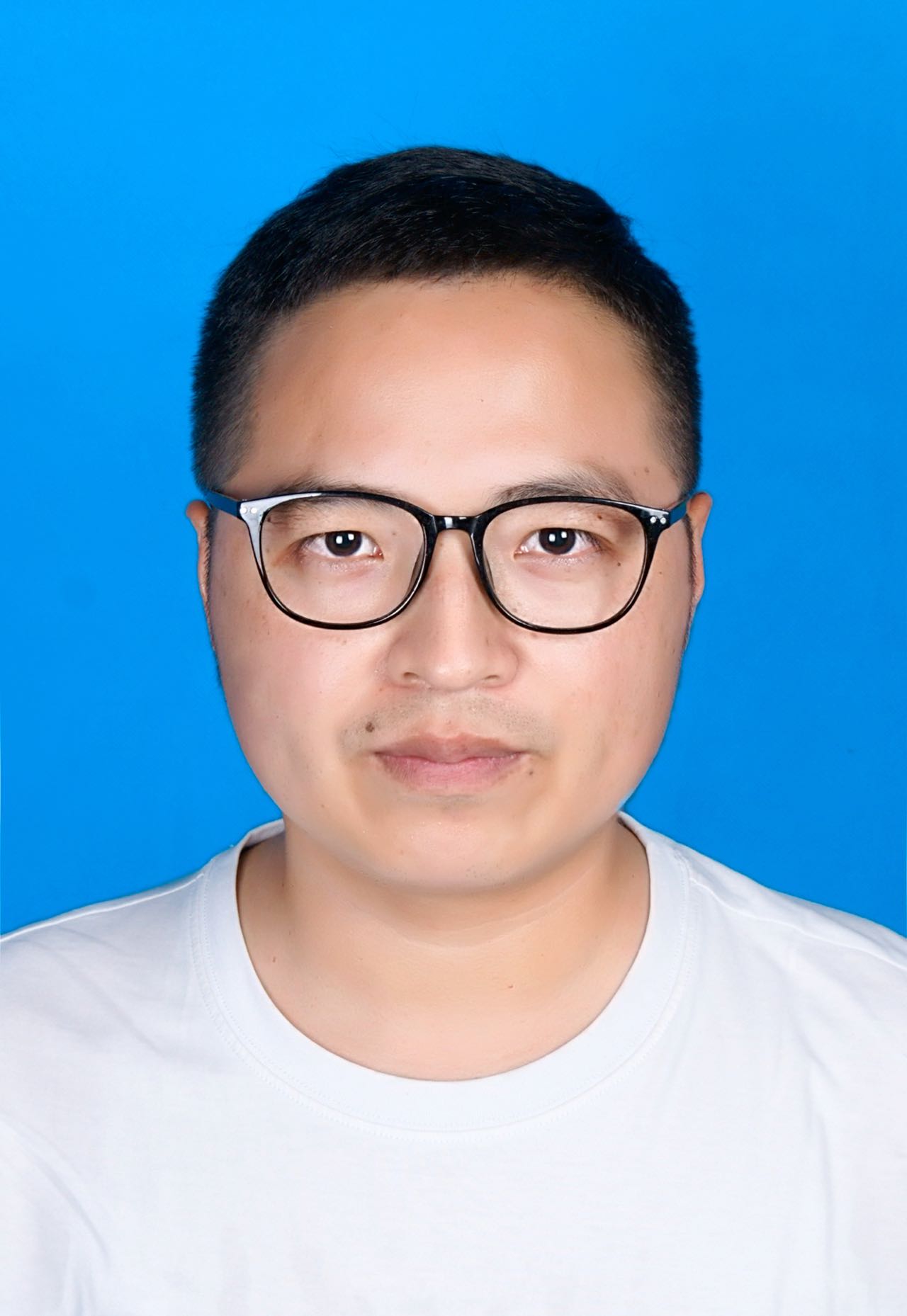}}]
{Zhuang Yang} received Ph.D. degree in School of Informatics from Xiamen University, Xiamen, China in 2018. He is currently an Associate Professor in the School of Computer Science and Technology, Soochow University, Suzhou, China. He was a Postdoctoral Fellow with Sun Yat-Sen University in 2019-2021 before joining Soochow University. His current research interests include machine learning, data science, computer vision, matrix analysis, optimization algorithm. His recent efforts mainly focus on the foundation and computation of high-performance algorithms for large-scale model.
\end{IEEEbiography}
\end{document}